\documentclass[11pt,a4paper]{article}
\usepackage[hyperref]{acl2017}
\usepackage{times}
\usepackage{graphicx}
\usepackage{latexsym}
\usepackage{gnuplot-lua-tikz}

\usepackage{url}

\aclfinalcopy 
\def\aclpaperid{181} 

\def\Tref#1{Table~\ref{#1}}
\def\Fref#1{Figure~\ref{#1}}
\def\Sref#1{Section~\ref{#1}}

\def\perscite#1{\citet{#1}}  
\def\inparcite#1{\citealp{#1}}  
\def\parcite#1{\cite{#1}}  

\title{Curriculum Learning and Minibatch Bucketing \\ in Neural Machine Translation}

\author{Tom Kocmi \and Ond{\v{r}}ej Bojar \\
  Charles University, \\
  Faculty of Mathematics and Physics \\
  Institute of Formal and Applied Linguistics \\
  {\tt {surname}@ufal.mff.cuni.cz} \\}

\date{}

\begin{document}
\maketitle
\begin{abstract}
We examine the effects of particular orderings of sentence pairs on the on-line
training of neural machine translation (NMT). We focus on two types of such
orderings: (1) ensuring that each minibatch contains sentences similar in some
aspect and (2) gradual inclusion of some sentence types as the training
progresses (so called ``curriculum learning''). In our English-to-Czech
experiments, the internal homogeneity
of minibatches has no effect on the training but some of our ``curricula'' achieve
a small improvement over the baseline.
\end{abstract}

\section{Introduction}

Machine translation (MT) has recently seen another major change of paradigms.
MT started with rule based approaches which worked successfully for small
domains. Generic MT was first reached with statistical methods, the early
word-based and the late phrase-based dominant approaches, that build upon large
training data. The current change is due to the first successful application of
deep-learning methods (neural networks) to the task, giving rise to neural MT
(NMT) \citep{collobert2011natural, sutskever2014sequence}. The data-driven methods have always been resource-heavy (e.g. word
alignment needing a day or two for large parallel corpora) and NMT pushed this
to new extremes: to reach the state-of-the-art performance, the model often
needs a few weeks on the highly parallel graphics processing units (GPUs),
equipped with large memory (8--12 GB) on a large training corpus.

The complexity of the training is a direct consequence of the complexity of the
neural MT model: we need to find optimal setting of dozens millions of
real-valued NMT model parameters that, according to the hard-coded model
structure, define the calculation that converts the sequence of source words to
the sequence of target words. The core of NMT training is thus numerical
optimization, gradient descent, towards the least error as defined by the
objective function. The common practice is to evaluate cross entropy against the
reference translation.

The gradient of the objective function, in which the algorithm progresses, can
be established on the whole dataset (called ``batch training''), on individual
examples (``online training'') or a small set of examples (``minibatch
training''). The full batch training has a clear advantage of reliable gradient
estimates, while online training can easily suffer from instability. As
documented by \perscite{wilson2003general} on 27 learning tasks,
online training reaches the same level of optima as the full batch training
while having
much lower memory demands and faster computation in
general.

Minibatches typically contain 50 to 200 examples, calculate and average the
error for all of them and propagate the error back through the network to update
the weights. They have the advantages of both: the gradient is more stable
and we decide how much of the training data it is convenient to handle at each
training step. A further benefit comes from parallelizability on GPUs: the error
of all the examples in the batch can be calculated simultaneously with the exact
same formulas.

The training sets in NMT are simply too large, so full batch training is out of
question and everybody uses minibatches.\footnote{In fact, the terms ``batch'' or batch size in NMT refer to
minibatches; the whole corpus is then called an ``epoch''.}
The benefit of parallelization in minibatches can be somewhat
diluted if minibatches contain sentences of varying length. In common frameworks
for parallel computation, all the items in the
minibatch must usually have the same length, and shorter sentences are therefore
padded with dummy symbols. Calculations over the padded areas are wasted.

\perscite{khomenko2016accelerating} and \perscite{doetsch2017comprehensive}
report improvements in training speed by organizing (bucketing) training sentences
so that sentences of identical or similar length arrive in the same minibatches.
A related idea is called ``curriculum learning'' \parcite{bengio2009curriculum}
where the network is first trained with easier examples, making the task more
complex only gradually.

In this work, we attempt to
improve the final
translation quality
and/or reduce the training time
of an NMT system by organizing minibatches in two particular
ways. In \Sref{bucketing}, minibatches are created to contain sentences similar
not only in length but in other (linguistic) phenomena, hoping for a better quality. In \Sref{curriculum},
similar criteria are used to organize the whole corpus,
increasing the complexity of examples as training progresses,
aiming at a better quality in shorter time.
\Sref{experiments} evaluates our ideas in thems of translation quality and
discusses the results. Related work is summarized in \Sref{related} and we
conclude
in \Sref{conclusion}.

\section{Minibatch Bucketing}
\label{bucketing}

Minibatches stabilize the online training from fluctuations \parcite{murata1999statistical}
and help to avoid a problem with overshooting local optima.

As mentioned, better performance of parallel processing has been achieved by
bucketing training examples to contain sentences of similar length. The benefit
of this approach however comes purely from the technical reason: avoiding
wasted computation on paddings.

Each minibatch leads to one update of the model parameters and each example in
the minibatch contributes to the average error. We assume that if all the
examples in the minibatch are similar in some \emph{linguistic sense}, they
could jointly highlight the fitness of the current model in this particular
aspect.
Each minibatch would be thus focused on some particular language phenomenon and
the gradient derived from this minibatch could improve the behavior of the
model in this respect, allowing the network an easier identification of shared
features of the examples.

%


We experiment with several features, by which we bucket the data. Those features
are: sentence length, number of coordinating conjunctions, number of nouns, number of proper
nouns
and the
number of verbs in the training data pairs. In our experiments we do not mix
features together, but such mixed-focus minibatches are surely also possible.

The exact procedure of training corpus composition is the following:
First, we divide all data based on their features into separate buckets (e.g. one
bucket of sentences with at most one verb, another bucket of sentences with two
or three verbs etc.). 
We then shuffle all examples in each bucket and break them down to groups of size same as the minibatch size.
Finally, all these groups are shuffled and concatenated. The corpus is then read
sequentially but our shuffling procedure ensured that all minibatches contain
data having the same feature but among minibatches, the features are shuffled.


\section{Curriculum Learning}
\label{curriculum}

When humans are trained, they start with easier tasks and gradually, as they
gain experience and abstraction, they are able to learn to handle more
and more complex situations. It has been shown by
\perscite{bengio2009curriculum}
that even neural networks can improve their performance when they are
presented with the easier examples first.

For neural networks, it is important to keep on training also on the easy examples,
because the networks are generally prone to very quick overfitting as we discuss in
\Sref{quick-overfit}.
If the network was presented only with the more difficult
examples, its performance on the easy ones would
drop. Some mixing strategy is thus needed.

%

\perscite{bengio2009curriculum} propose a relatively simple strategy. They
organize all training data into bins of similar complexity. The training then
starts with all the examples in the easiest bin (step-by-step in minibatches).
With the easiest bin covered, the first and second easiest bins are allowed.
In the final stage, examples from all the bins are used in the training.

The disadvantage of this approach is that examples in easier batches are
processed several times. This boosts their importance for the training and also
prevents us from directly comparing this strategy with the baseline of simply
shuffled corpus.


We improve this strategy to use each example only once during an epoch.
For our method to work, we require that the number of examples in the
bin only decreases as we move to the bin of the higher complexity. This is
usually easy to reach as there are generally more easier sentence pairs than
complex sentence pairs in parallel corpora. The bin thresholds can be also
adjusted
to fulfill this condition.

The strategy for selecting examples from the bins is the following. First, we draw
examples from the easiest bin only until there remain the same number of
examples
as in the second most easy bin. We then continue to draw uniformly from the
first two easiest bins until in each of them, there remain the same number of
examples as in the third one, etc. When taking the examples, we always
accumulate one minibatch and feed it to the training. If the number of bins is
smaller than the size of the minibatch, the minibatches in the late stages will
contain examples from all complexity bins. If there are more bins than the
minibatch size, each minibatch will be highly varied in complexity and the
training will gradually proceed over examples of all complexities.


\subsection{Selected Features}

It is not entirely clear which examples are easy and which are hard for NMT (in
various stages of the training).
We experiment with several linguistically-motivated features.

The first feature is the length of the target sentence. (Source sentences
usually have a corresponding length.) Our bins are for sentences of up to 8
tokens, up to 12 tokens, 16, 20, up to 40 tokens and for longer sentences.
The thresholds were chosen to satisfy the requirement of more examples in easier
bins.

The second binning is based on the number of coordinating conjunctions in the
target sentence as one possible (rough) estimate of the number of clauses in the
sentence. Conjuctions are also used in lists of items, so a higher number of
them suggests that the sentence structure is cluttered with lists. Such examples
may be easy to translate but do not correspond well to the generally
hierarchical structure of sentences that we want to expose to the network.
We use the same thresholds as for sentence length.

Learners of foreign languages often read books written with a simplified
vocabulary. To replicate this learning strategy, we sort words by their
decreasing frequency and define ranks on this list. For example, the first rank
contains the 5000 most frequent words. Sentences are then organized into bins
based on the least frequent word in them: the first bin contains sentences with
all the words appearing the first rank.

We define the ranks separately for source and for target language and experiment
with binning based on one of them or both at the same time.

\section{Experiments}
\label{experiments}

This section describes our experiments and results with minibatch bucketing and
curriculum learning.

\subsection{Model Details}

We use Neural Monkey \parcite{NeuralMonkey:2017}, an open-source neural machine
translation and general sequence-to-sequence learning system built using the
TensorFlow machine learning library.

Neural Monkey is quite flexible in model configuration but we restrict our
experiments to the standard encoder-decoder architecture with attention as
proposed by \perscite{bahdanau:2014:corr}.
We use the same model parameters as defined for the WMT 2017 NMT Training
Task \citep{trainingtask2017}.
The task
defines models of two sizes, one that fits a 4GB GPU and one that fits an 8GB
GPU. We use the former one where the encoder uses embeddings of size 300 and the
hidden state of 350. Dropout is turned off and maximum input sentence length is
set to 50
tokens. The decoder uses attention mechanism and conditional GRU cells, with the
hidden state of 350. Output embedding has the size of 300, dropout is turned off
as well and the maximum output length is again 50 tokens.
The Adam \cite{journals-corr-KingmaB14} optimizer is used as the gradient descend algorithm.

To reduce vocabulary size, we use byte pair encoding
\parcite{sennrich-haddow-birch1} which breaks all words into subword units
defined in the vocabulary. The vocabulary is initialized with all letters and
larger units are added on the basis of corpus statistics. Frequent words make it
to the vocabulary, less frequent words are (deterministically) broken into
smaller units from the vocabulary.

As defined for the NMT Training Task, we set the vocabulary of size to 30,000
subword units. The vocabulary is constructed jointly for the source and target
side of the corpus.

During the inference, we use simple greedy algorithm which generates the most frequent
word depending on the previously generated words, the state of the decoder and
attention. We did not employ any better decoding algorithm such as beam search \parcite{beam1, beam2}
mainly due to technical difficulties. Although this decision leads to a poorer performance,
it should not have any influence on the results of our work.

All experiments are based on one epoch of training over whole training dataset. The
training takes roughly one week on NVIDIA GeForce GTX 1080. We
should note that our model used only 4 GB of memory, instead of 8 GB available
in the GPUs.

For the plots and presentation of the results, we compute test score (BLEU,
\inparcite{papineni:2002}) after
every 100k training examples. 
To compensate for fluctuations
during the training, we report the mean and standard
deviation of the last 10
test errors of the training. This simple smoothing method is a substitute for
proper significance testing \parcite{clark-EtAl:2011},
since we cannot run all experiments
multiple times due to the lack of computing resources.

\subsection{Training Data}

We use the dataset provided for the WMT 2017 NMT Training Task. The dataset
comes from the CzEng 1.6 corpus \parcite{bojar2016czeng} and it was cleaned by
the organizers of the NMT Training Task. The resulting corpus is 48.6 million
sentence pairs for English-to-Czech translation.

We use the test set from the WMT 2016 News Translation Task as our only heldout
set. We do not need any separate development or validation set, because we are
not doing any hyperparameter search or run experiments several times to find the
best-performing setup.

\begin{table}[t]
\begin{center}
\small
\begin{tabular}{lcc}
Feature         & Performance score \\
\hline
None (baseline)  & 14.25 $\pm$ 0.18 BLEU\\
\hline
Number of conjuctions   & 14.71 $\pm$ 0.24 BLEU\\
Number of proper nouns  & 14.58 $\pm$ 0.22 BLEU\\
Number of nouns  & 14.57 $\pm$ 0.24 BLEU\\
Sentence length  & 14.43 $\pm$ 0.23 BLEU\\
Number of verbs  & 14.43 $\pm$ 0.21 BLEU\\
\end{tabular}
\end{center}
\caption{Minibatch bucketing after one epoch.}
\label{bucketing-results}
\end{table}

\subsection{Minibatch Bucketing}

\Tref{bucketing-results} shows the results of our experiments with minibatch
bucketing. The bucketed runs are slightly better than the baseline but they
usually fall in the standard deviation range so we cannot claim any significant
improvement.

\subsection{Curriculum Learning}

This sections describes our experiments with curriculum learning. We 
organized the training data based on the following features: the length of the sentences,
the number of coordinating conjunctions, the highest rank of a word in the
Czech or the English part and two combinations of the word ranks: ``max word
rank'' which puts sentences into bins based on the maximum rank of their English
and Czech words and ``combined rank'' is based on word ranks derived from
concatenated source and target corpora.


As documented in \Tref{curriculum-results}, several of the curriculum setups
improve over the baseline. The most beneficial is to organize the bins by the
(target-side) sentence length, reaching a gain of 1 BLEU point.

\begin{figure*}
\begin{center}
\input{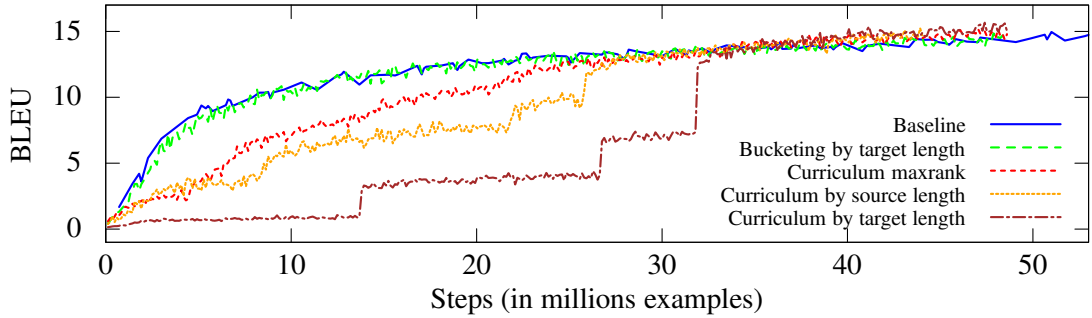}
\end{center}
\caption{Selected learning curves for minibatch bucketing and curriculum.}
\label{strategies}
\end{figure*}

\begin{table}[t]
\begin{center}
\small
\begin{tabular}{lc}
Feature         & Performance score \\
\hline
None (baseline)  & 14.25 $\pm$ 0.18 BLEU\\
\hline
Source sentence length  & 15.41 $\pm$ 0.18 BLEU\\
Target sentence length  & 15.24 $\pm$ 0.27 BLEU\\
English word ranks   & 15.07 $\pm$ 0.28 BLEU\\
Czech word ranks  & 15.06 $\pm$ 0.29 BLEU\\
Number of conjuctions  & 15.04 $\pm$ 0.24 BLEU\\
Combined word ranks   & 14.77 $\pm$ 0.16 BLEU\\
Max word ranks   & 14.73 $\pm$ 0.22 BLEU \\
\end{tabular}
\end{center}
\caption{Curriculum learning after one epoch.}
\label{curriculum-results}
\end{table}

\Fref{strategies} plots learning curves for the baseline, one
minibatch bucketing run (\Sref{bucketing}) and some curricula setups.
Bucketing closely follows the baseline while curricula start much worse and make
up later, as the complexity of training examples matches the fixed complexity of
the test set.

The difference between source- and target-length curriculum is particularly
interesting. Binning by target length ensures strict target-sentence limits and
the decoder indeed follows the restriction never producing longer sentences
regardless the source length. This results in serious penalization, see the sharp jumps in
``Curriculum by target length''.
Source-side binning makes
target lengths slightly more varied.
Assuming some model of sentence length in the decoder
\parcite{shi-knight-yuret:etal:2016},
training it on strictly capped sentences seems to damage its learning while the
more varied data better allow to learn to
predict output length based on the input length.


\subsection{Quick Adaptation or Overfitting}
\label{quick-overfit}

\begin{figure*}
\begin{center}
\input{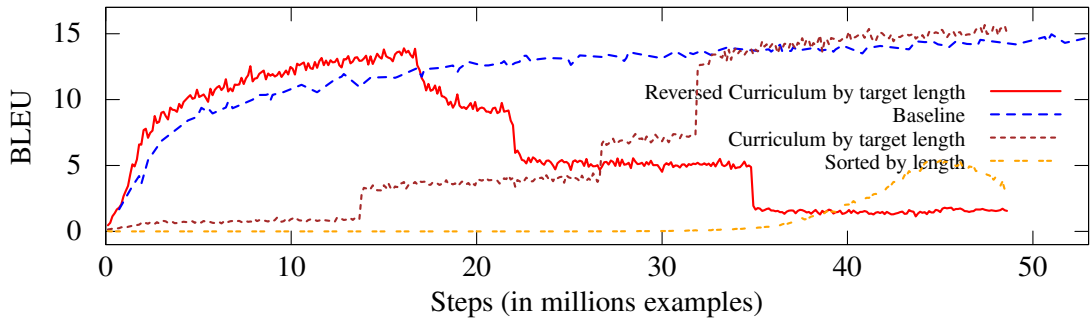}
\end{center}
\caption{Learning curves of selected curriculum learning runs and other
contrastive runs.}
\label{learningcurves}
\end{figure*}

Neural networks are known to  quickly adapt to new types of data as they arrive
in the training.
This effect is used e.g. in domain adaptation for NMT
\citep{freitag2016fast,luong2015stanford}
but there is a big risk of 
overfitting to some specialized data.

As shown in \Fref{learningcurves}, our curriculum runs are heavily affected by
this quick adaptation. ``Baseline'' shows the standard behaviour: starting
quickly and then more or less flattening
towards the end of the epoch.

Our best performing curriculum setup starts with
short sentences and the model thus first learns to produce only short sentences. The
curve ``Curriculum by target length'' shows very bad scores for more than a half
of the training data, and the particularly striking are the quick transitions
whenever a new bin of longer sentences is added. The model adapts and starts
producing longer sentences, getting a huge boost in BLEU on the fixed test set.
Towards the end of the epoch, ``Curriculum by target length'' demonstrates
its improved generalization power and surpasses the baseline.

If we did not use our strategy of revisiting shorter sentences and simply sorted
the corpus by sentence length, the training would fail spectacularly, see the
curve ``Sorted by length''. The model never reaches any reasonable performance.

The curve ``Reversed Curriculum by target length'' is very interesting. We
simply took the best corpus organization (``Curriculum by target length'') and reversed
it. The training performs better in the early stages (i.e. minibatches evenly
covering all length bins) but very quickly drops as the long-sentence bins get
prohibited. Put differently, the model quickly adapts (overfits) to the new
``domain'' of short sentences and fails to produces normal-length translations
of the test set.

\subsection{Continuing the Curriculum}

\begin{figure*}
\begin{center}
\input{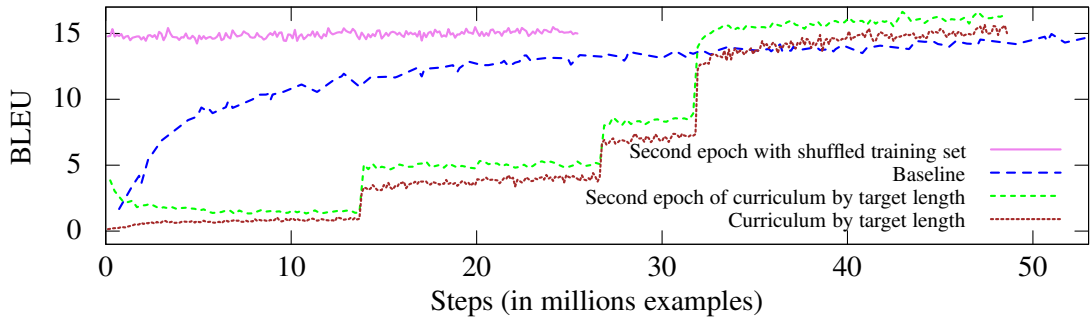}
\end{center}
\caption{The best run (``Curriculum by target length'') and its continuations.}
\label{continuation}
\end{figure*}

It should be noted that all  results presented so far are observed after one 
 epoch of curriculum training.
It is questionable what would be the best way of subsequent training.

We considered two options, see \Fref{continuation}. Starting
over from the easiest examples harms the performance terribly early in the epoch
but succeeds in improving the performance of the first epoch all the time, see the ``Second
epoch of curriculum by target length'' in \Fref{continuation}.

Another option is
to continue the
training after the first epoch with the training dataset shuffled. As the
corresponding curve in \Fref{continuation} however shows, the model is probably
already quite fixed in the current optimum and we do not see any further improvement on
the test set.

\section{Related Work}
\label{related}

\perscite{khomenko2016accelerating} used a bucketing technique to accelerate the
speed of the training. They prepared minibatches of training data with
similar length and got a speedup in the training time of factor up to 4. The
buckets are drawn randomly from the training set. A similar approach is also
used in Nematus \parcite{sennrich-EtAl:2017:EACLDemo}, one of the
state-of-the-art open-source toolkits for NMT.

\perscite{doetsch2017comprehensive} used bucketing and experimented with
ordering of the bucketed batches. Their proposed method orders buckets in an
alternating way: first in increasing order by length,
then decreasing order, then again increasing order etc. This way the buckets of
different length are periodically revisited. With this approach, the authors got
a speedup in the training time and also obtained better performance results.

\perscite{bengio2009curriculum} use curriculum learning 
for a neural language model, not a full NMT system. They trained the
network by iteratively
increasing the vocabulary size, starting with the vocabulary of 5000 and increasing by
5000 each epoch. Each epoch used only sentences with words available in the
current restricted vocabulary.
The last epoch thus used
all examples. This curriculum lead to a statistically significant improvement in
the performance of the model.

\perscite{graves2017automated} automatically
select  examples  during multitask learning. The method evaluates
training signals from the neural network and uses them to focus on
specific subtasks to
accelerate the training process of the main task. The authors noted that uniformly sampling from the 
training data is a strong baseline.

\section{Conclusion}
\label{conclusion}

We examined the effects of two ways of orderings of training examples for neural
machine translation from English to Czech.

Trying to use sentences with similar linguistic properties in each minibatch of
the online training (dubbed ``minibatch bucketing'') did not bring any
difference from the baseline of randomly composed minibatches.

Organizing minibatches to gradually include more complex sentences (in terms of
length or vocabulary size) helps to reach better translation quality of up to 1
BLEU point.

The actual process of learning is however very interesting,
displaying clear jumps in the performance as longer sentences are added to the
training data.
The strategy cannot be thus used to shorten the training time: unless the
gradually-organized epoch is finished, the model performs well below the
baseline.

Our experiments also confirm the quick adaptability of deep learning
methods, with a high risk of overfitting to particular properties of the
very recent training examples.

\section*{Acknowledgement}

This study was partly supported by the grants SVV~260~453, GAUK 8502/2016, H2020-ICT-2014-1-645442 (QT21) and Charles University Research Programme ``Progres'' Q18 -- Social Sciences: From Multidisciplinarity to Interdisciplinarity. It has been using language resources and tools from LINDAT/CLARIN project of the Ministry of Education, Youth and Sports of the Czech Republic (project LM2015071).
 
%

\bibliography{biblio}
\bibliographystyle{acl_natbib}


\end{document}